\documentclass{article}

\PassOptionsToPackage{numbers, compress}{natbib}

\usepackage[preprint]{neurips_2022}

\usepackage[utf8]{inputenc} 
\usepackage[T1]{fontenc}    
\usepackage[colorlinks=true,citecolor=blue]{hyperref}%
\usepackage{url}            
\usepackage{booktabs}       
\usepackage{amsfonts}       
\usepackage{nicefrac}       
\usepackage{microtype}      
\usepackage{xcolor}         
\usepackage{xcolor}         
\usepackage{lipsum}
\usepackage{afterpage}
\usepackage{comment}
\usepackage{makecell}

\usepackage{soul}
\sethlcolor{white} 
\DeclareRobustCommand{\hln}[1]{{\sethlcolor{white}\hl{#1}}}

\title{\hl{Distinction Maximization Loss: Efficiently Improving Out-of-Distribution Detection and Uncertainty Estimation by Replacing the Loss and Calibrating}}

\author{%
David~Mac\^edo\textsuperscript{1,2}\blfootnote{did},~Cleber~Zanchettin\textsuperscript{1},~Teresa~Ludermir\textsuperscript{1}
\\
\textsuperscript{1}Centro de Inform\'atica, Universidade Federal de Pernambuco, Brasil\\
\textsuperscript{2}Montreal Institute for Learning Algorithms, University of Montreal, Canada\\
\texttt{dlm@cin.ufpe.br,\,cz@cin.ufpe.br,\,tbl@cin.ufpe.br}\\
}

\usepackage{graphicx}
\usepackage{footmisc}
\usepackage{booktabs}
\usepackage{amsmath}
\usepackage{amsfonts}
\usepackage{amsbsy}
\usepackage{placeins}
\usepackage{siunitx}
\usepackage{bm}
\usepackage{multirow}
\usepackage{mathtools}
\usepackage{xparse}
\DeclarePairedDelimiter\abs{\lvert}{\rvert}
\DeclarePairedDelimiter\norm{\lVert}{\rVert}

\usepackage{empheq}
\usepackage{tabularx} 
\newcolumntype{C}{>{\centering\arraybackslash}X}
\usepackage{subfig}
\usepackage{centernot}
\usepackage{ragged2e}

\newcommand\blfootnote[1]{%
  \begingroup
  \renewcommand\thefootnote{}\footnote{#1}%
  \addtocounter{footnote}{-1}%
  \endgroup
}

\usepackage{tikz}
\usetikzlibrary{backgrounds}
\usetikzlibrary{arrows,shapes}
\usetikzlibrary{tikzmark}
\usetikzlibrary{calc}
\usepackage{amsmath}
\usepackage{amsthm}
\usepackage{amssymb}
\usepackage{mathtools, nccmath}
\usepackage{wrapfig}

\usepackage{blindtext}

\usepackage{graphicx}

\usepackage{xspace}

\usepackage{array}
\usepackage{ragged2e}
\newcolumntype{P}[1]{>{\RaggedRight\hspace{0pt}}p{#1}}

\usepackage{tcolorbox}

\usepackage{tikz}
\usetikzlibrary{arrows,shapes,positioning,shadows,trees,mindmap}
\usepackage[edges]{forest}
\usetikzlibrary{arrows.meta}
\colorlet{linecol}{black!75}
\usepackage{xkcdcolors} 

\usepackage{tikz}
\usetikzlibrary{backgrounds}
\usetikzlibrary{arrows,shapes}
\usetikzlibrary{tikzmark}
\usetikzlibrary{calc}
\newcommand{\highlight}[2]{\colorbox{#1!17}{$\displaystyle #2$}}

\renewcommand{\highlight}[2]{\colorbox{#1!17}{#2}}

\usepackage{tikz}
\usepackage{pgfplots}
\usepackage{tikz-3dplot}
\tdplotsetmaincoords{60}{115}
\pgfplotsset{compat=newest}

\DeclarePairedDelimiterX{\infdivx}[2]{(}{)}{%
  #1\;\delimsize|\delimsize|\;#2%
}
\newcommand{\kld}[2]{\ensuremath{D_{KL}\infdivx{#1}{#2}}\xspace}

\begin{document}

\maketitle

\begin{abstract}
Building robust deterministic neural networks remains a challenge. On the one hand, some approaches improve out-of-distribution detection at the cost of reducing classification accuracy in some situations. On the other hand, some methods simultaneously increase classification accuracy, uncertainty estimation, and out-of-distribution detection at the expense of reducing the inference efficiency. In this paper, we propose training deterministic neural networks using our DisMax loss, which works as a drop-in replacement for the usual SoftMax loss (i.e., the combination of the linear output layer, the SoftMax activation, and the cross-entropy loss). Starting from the IsoMax+ loss, we create each logit based on the distances to all prototypes, rather than just the one associated with the correct class. We also introduce a mechanism to combine images to construct what we call fractional probability regularization. Moreover, we present a fast way to calibrate the network after training. Finally, we propose a composite score to perform out-of-distribution detection. Our experiments show that DisMax usually outperforms current approaches simultaneously in classification accuracy, uncertainty estimation, and out-of-distribution detection while maintaining deterministic neural network inference efficiency. The code to reproduce the results is available.
\footnote{\url{https://github.com/dlmacedo/distinction-maximization-loss}}
\end{abstract}

\section{Introduction}

Deep neural networks have been used for classification in many applications. However, improving the robustness of such systems remains a significant challenge. Classification accuracy, uncertainty estimation, and out-of-distribution (OOD) detection comprise three essential points regarding measuring the robustness of deep learning approaches.

On the one hand, some OOD detection approaches do not address uncertainty estimation or produce diminished classification accuracy in some cases \cite{techapanurak2019hyperparameterfree, Hsu2020GeneralizedOD}. These solutions also require changing the training of the last layer by removing its weight decay to work correctly. Therefore, they do not work as straightforward drop-in replacements for the SoftMax loss (i.e., the combination of the linear output layer, the SoftMax activation, and the cross-entropy loss \cite{liu2016large}). On the other hand, some recent approaches that address both OOD detection and uncertainty estimation require hyperparameter tuning and reduce the inference efficiency compared to pure deterministic neural networks and do not increase classification accuracy \cite{lakshminarayanan2017simple, Amersfoort2020SimpleAS, DBLP:conf/nips/LiuLPTBL20}. 

Therefore, simultaneously increasing classification accuracy, OOD detection, and uncertainty estimation performances while maintaining inference efficiency poses a challenge, mainly if we also desire to avoid training the same architecture many times to tune hyperparameters.

Recently, so-called IsoMax loss variants have been proposed \cite{macdo2019isotropic, DBLP:journals/corr/abs-2006.04005, macedo2021enhanced}. They increase the OOD detection performance without reducing the inference efficiency compared to pure deterministic deep neural networks trained using the usual SoftMax loss. However, they increase neither the classification accuracy nor uncertainty estimation.

\begin{figure*}
\centering
\subfloat[]{\includegraphics[height=3.4cm]{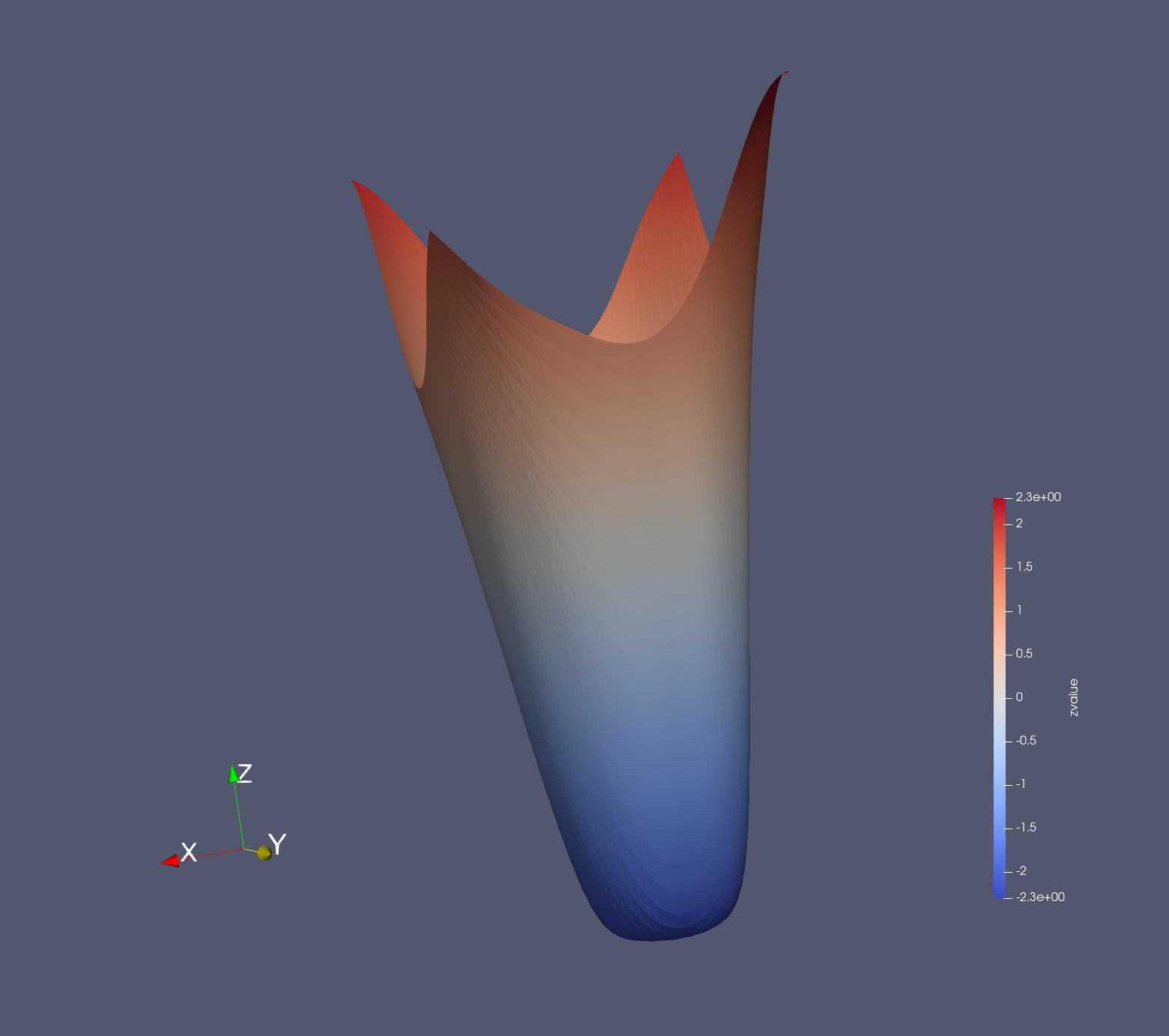}}
\hspace{0.05cm}
\subfloat[]{\includegraphics[height=3.4cm]{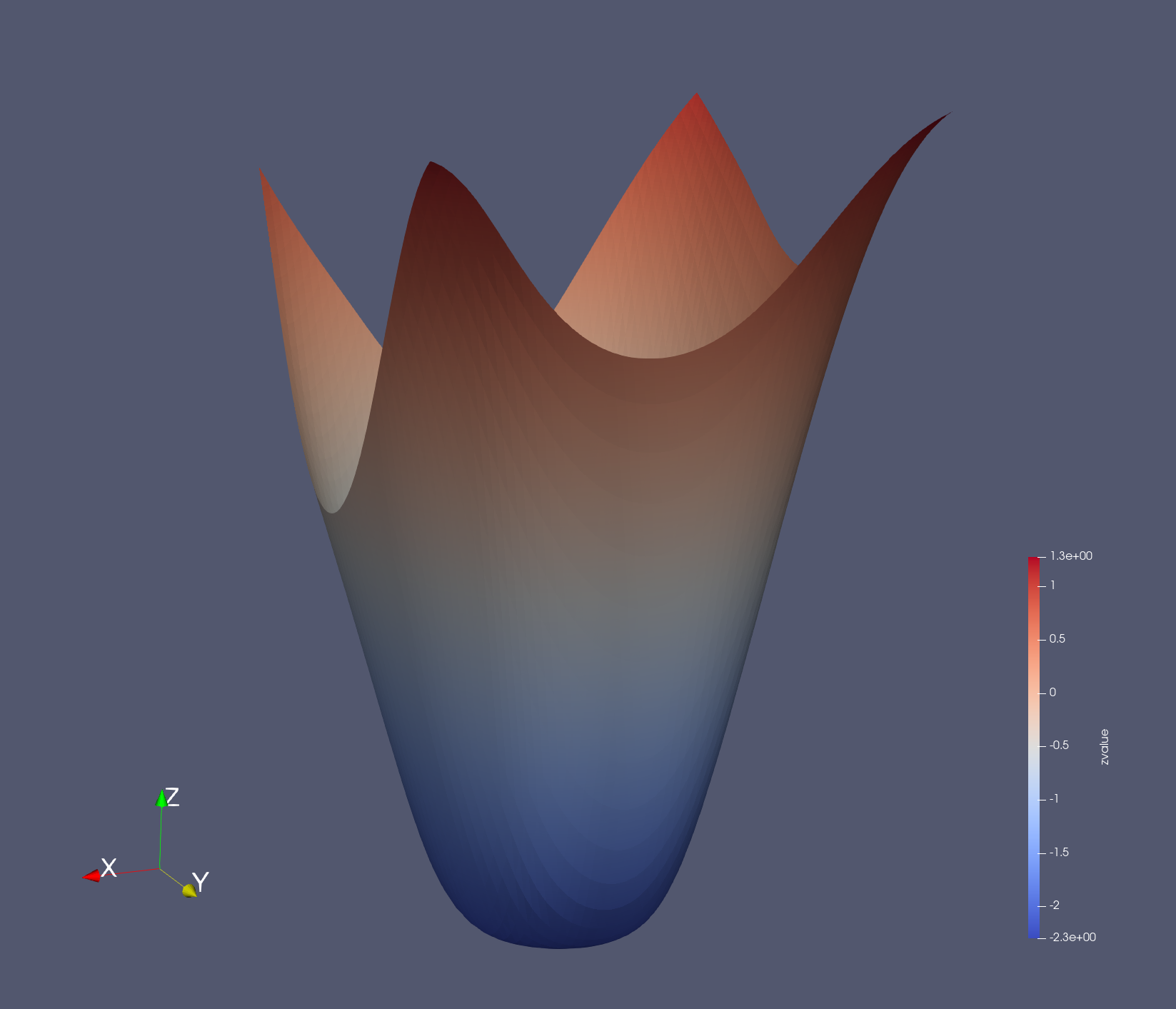}}
\hspace{0.05cm}
\subfloat[]{\includegraphics[height=3.4cm]{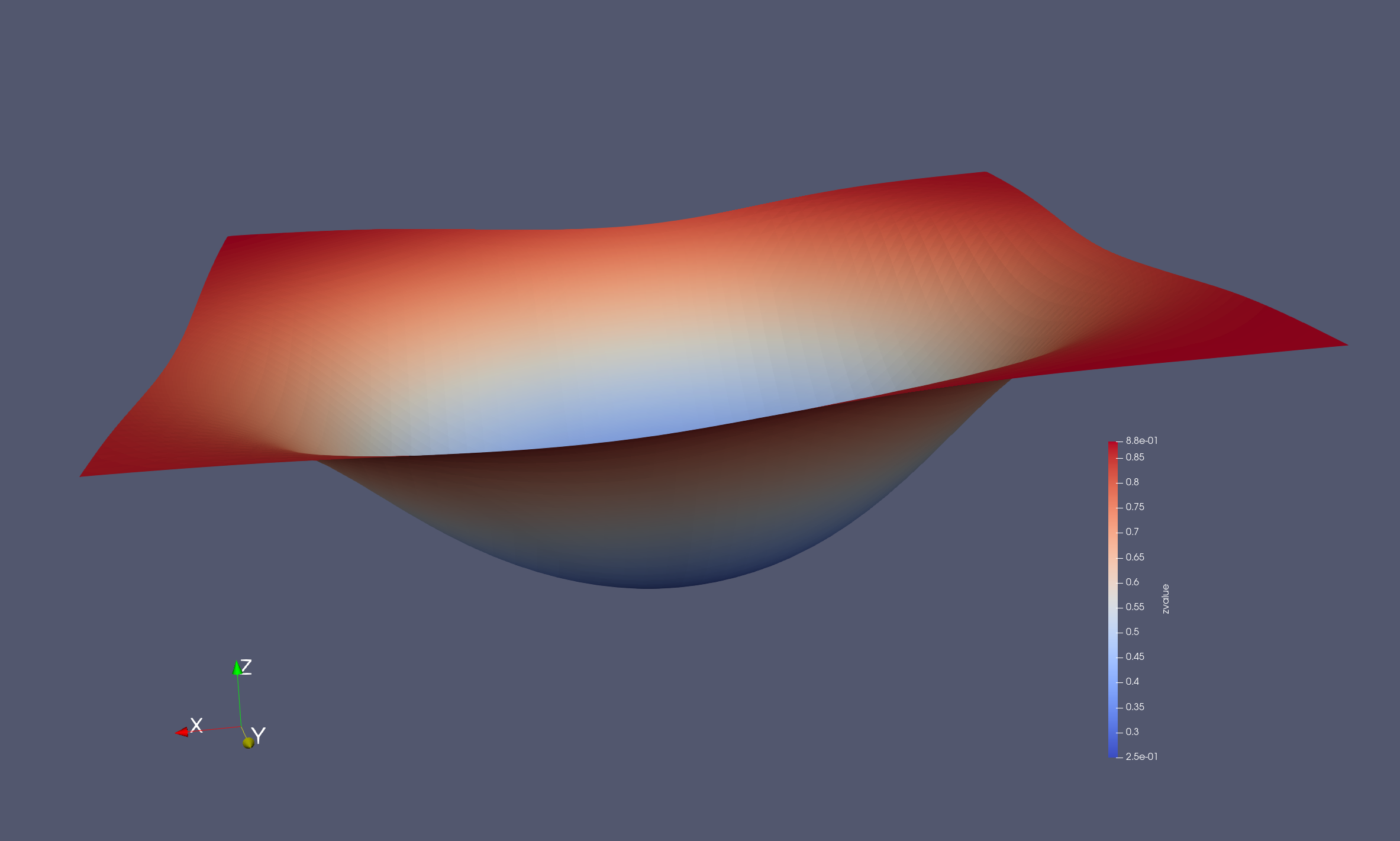}}
\\
\subfloat[]{\includegraphics[width=0.33\textwidth]{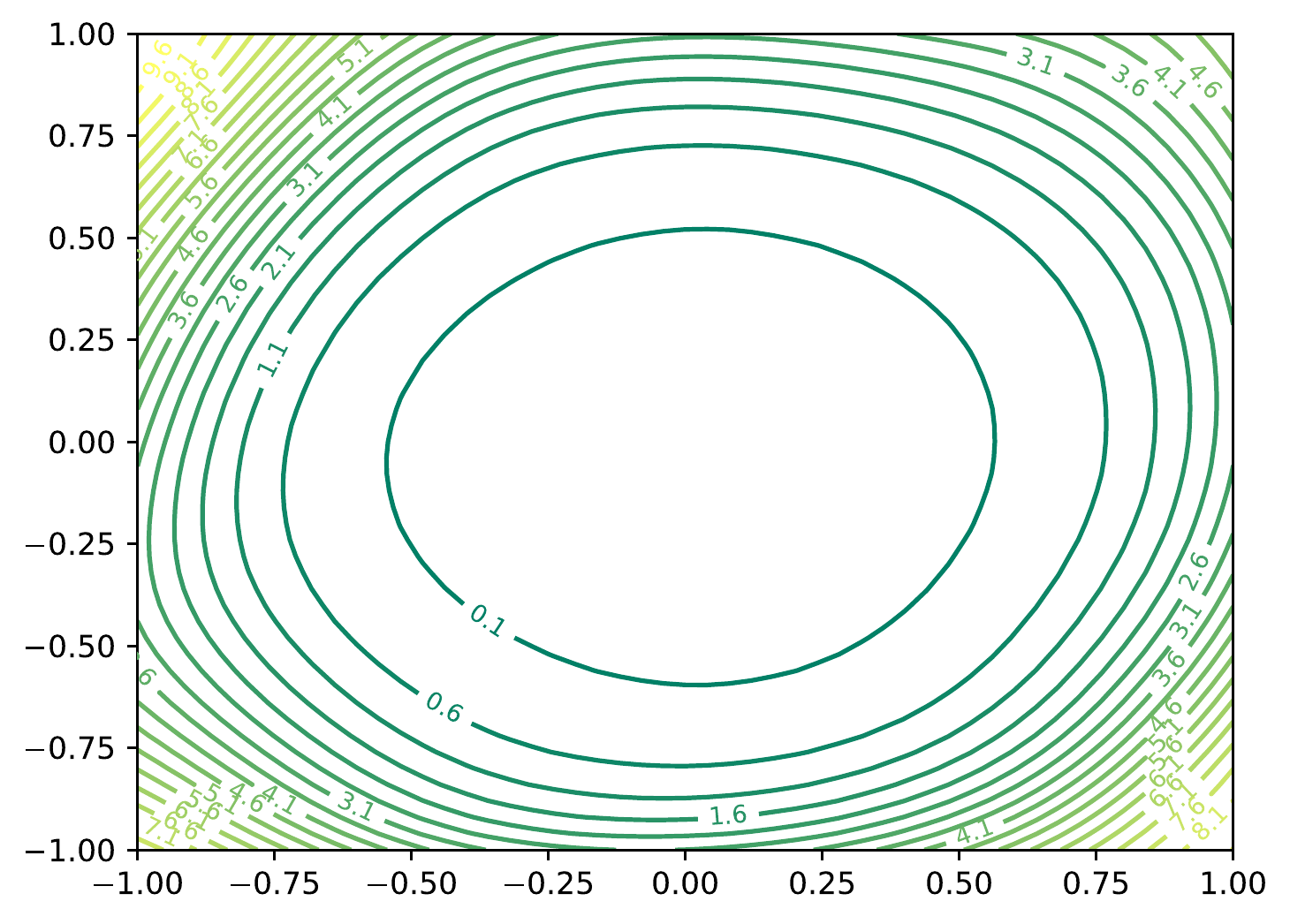}}
\subfloat[]{\includegraphics[width=0.33\textwidth]{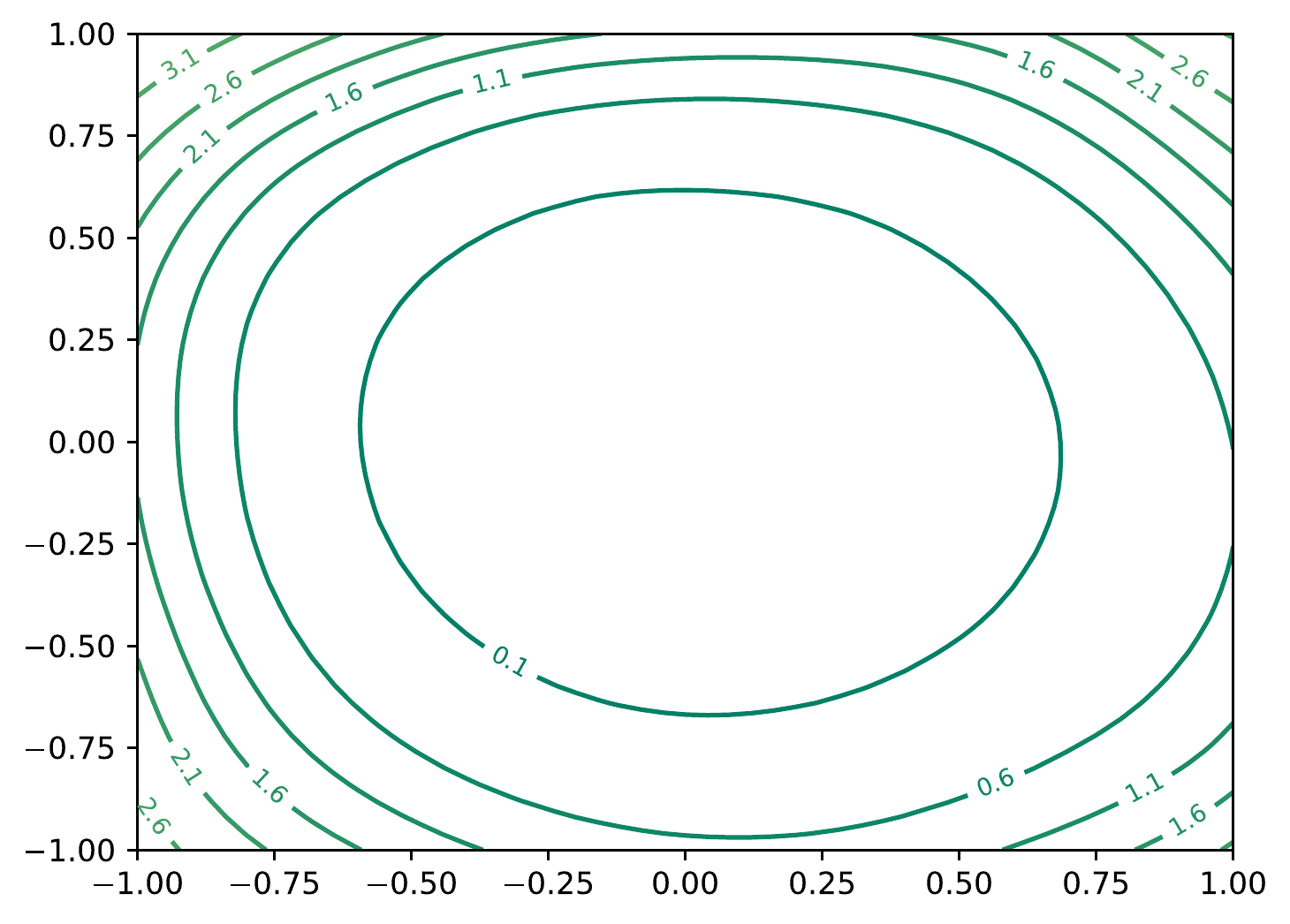}}
\subfloat[]{\includegraphics[width=0.33\textwidth]{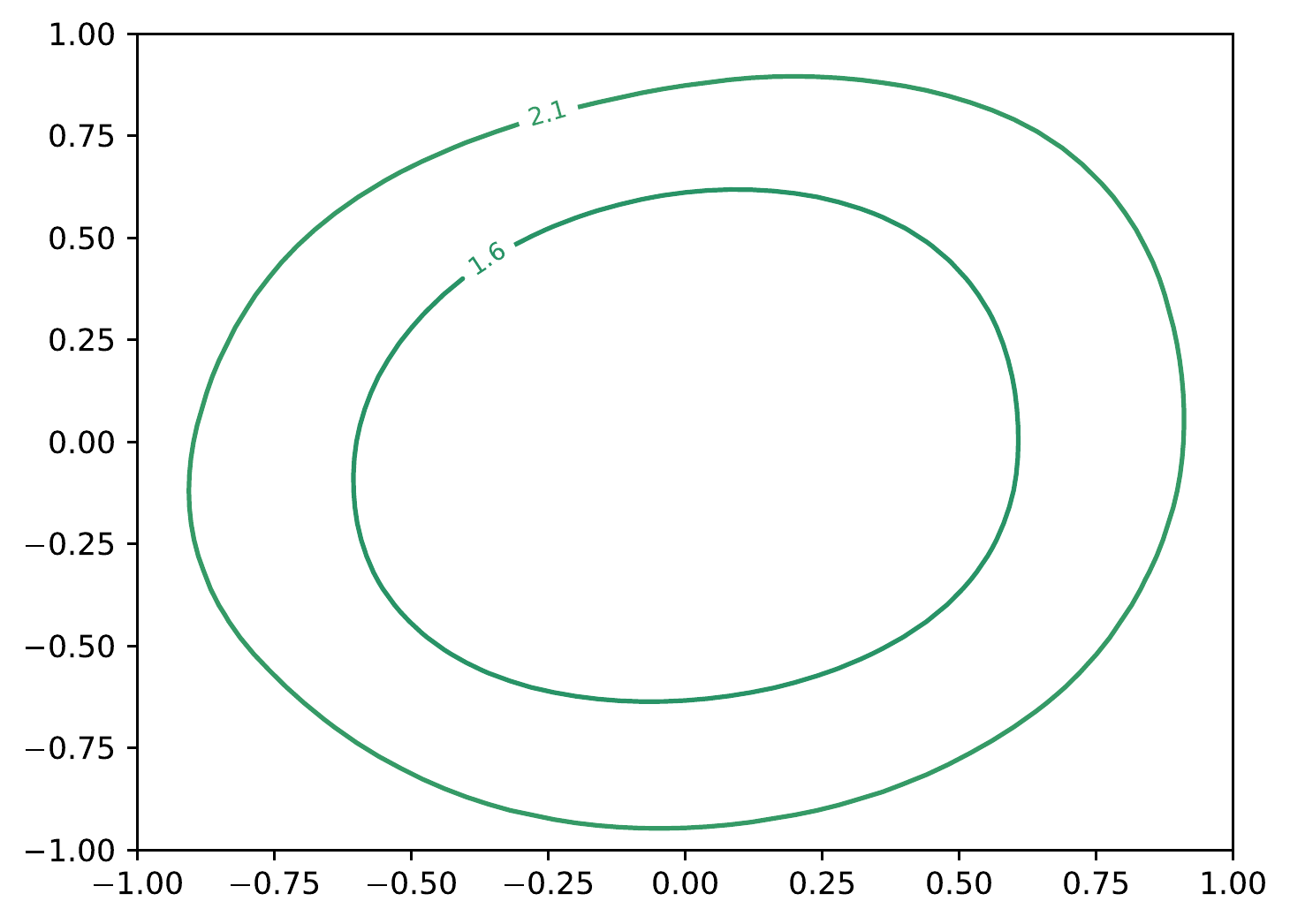}}
\caption{\textbf{Loss Surface Study.} 3D loss surfaces and 2D loss contours as proposed in \cite{DBLP:conf/nips/Li0TSG18}. Loss landscapes for ResNet34 trained on CIFAR10. (a, d) SoftMax; (b, e) IsoMax+; and (c, f) DisMax\textsuperscript{$\dagger$}. Considering that IsoMax+ outperforms SoftMax and DisMax\textsuperscript{$\dagger$} outperforms IsoMax+, a less steep 3D inclination (i.e. a lower 2D contour concentration) provides increased robustness.}
\label{fig:loss_landscape}
\end{figure*}

\paragraph{Contributions}In this paper, starting from IsoMax+ loss \cite{macedo2021enhanced}, we construct the Distinction Maximization (DisMax) loss. Our main contributions are the following. First, we create the \emph{enhanced} logits (logits+) by using \emph{all} feature-prototype distances, rather than just the feature-prototype distance to the correct class. \hl{The feature-prototype distances are the set of distances between a given feature and a given prototype.} Second, we introduce the \emph{fractional} probability regularization (FPR) by minimizing the Kullback–Leibler (KL) divergence between the output probability distribution associated with a \emph{compound} image and a target probability distribution containing fractional rather than integer probabilities.  \hl{The idea is to force the neural network to present outputs with higher entropies as preconized by the maximum entropy principle.} We call DisMax dagger (DisMax\textsuperscript{$\dagger$}) the variant of our loss when using FPR. Third, we construct a \emph{composite} score for OOD detection that combines three components: the maximum logit+, the \emph{mean} logit+, and the entropy of the network output. Fourth, we present a simple and fast temperature-scaling procedure that allows DisMax trained models to produce a high-performance uncertainty estimation. Like IsoMax+, DisMax works as a drop-in replacement for SoftMax loss. Moreover, when using DisMax, only a \emph{single} neural network training is required to use the proposed solution, as it avoids hyperparameter tuning. Furthermore, the trained models keep deterministic neural network inference efficiency. Finally, we show experimentally that to obtain improved robustness, we need to construct losses with less steep 3D landscapes, as showed and explained in Fig.~\ref{fig:loss_landscape}.

\newpage\section{Distinction Maximization Loss}\label{sec:dismax_loss}

\begin{figure}
\centering
\small
\subfloat[]{
\begin{tikzpicture}[tdplot_main_coords, scale = 0.8]
\node[align=center] at (0,0,3.6) {\color{black} \textit{(n-1)-sphere of radius one in the}\\\textit{IsoMax+ n-dimensional Euclidean space}};
\coordinate (P1) at ({1.75/sqrt(2)},{-1/sqrt(2)},{1.5/sqrt(2)});
\coordinate (P2) at ({(-1/sqrt(2))/2},{(1/sqrt(2))/2},{2/sqrt(2)});
\coordinate (P3) at ({2.5/sqrt(2)},{1.75/sqrt(2)},{0});
\coordinate (F1) at ({1.3/sqrt(2)},{2.5/sqrt(2)},{2/sqrt(2)});
\shade[ball color = lightgray, opacity = 0.5] (0,0,0) circle (2.0cm);
\tdplotsetrotatedcoords{0}{0}{0};
\draw[dotted, tdplot_rotated_coords, gray] (0,0,0) circle (2);
\tdplotsetrotatedcoords{90}{90}{90};
\draw[dotted, tdplot_rotated_coords, gray] (2,0,0) arc (0:180:2);
\tdplotsetrotatedcoords{0}{90}{90};
\draw[dotted, tdplot_rotated_coords, gray] (2,0,0) arc (0:180:2);
\draw[dotted, gray] (0,0,0) -- (-2,0,0);
\draw[dotted, gray] (0,0,0) -- (0,-2,0);
\draw[dotted, gray] (0,0,0) -- (0,0,-2);
\draw[-stealth] (0,0,0) -- (3.60,0,0) node[below left] {$x$};
\draw[-stealth] (0,0,0) -- (0,2.50,0) node[below right] {$y$};
\draw[-stealth] (0,0,0) -- (0,0,2.70) node[above] {$z$};
\draw[thick, -stealth] (0,0,0) -- (P1) node[above] {$P_1$};
\draw[thick, -stealth] (0,0,0) -- (P2) node[above] {$P_2$};
\draw[thick, -stealth] (0,0,0) -- (P3) node[below] {$P_3$};
\draw[thick, -stealth, blue] (0,0,0) -- (F1) node[above right] {$F$};
\draw[ultra thick, dashed, olive] (F1) -- (P1);
\end{tikzpicture}
}
\hskip 0.5 cm
\subfloat[]{
\begin{tikzpicture}[tdplot_main_coords, scale = 0.8]
\node[align=center] at (0,0,3.6) {\color{black} \textit{(n-1)-sphere of radius one in the}\\\textit{DisMax n-dimensional Euclidean space}};
\coordinate (P1) at ({1.75/sqrt(2)},{-1/sqrt(2)},{1.5/sqrt(2)});
\coordinate (P2) at ({(-1/sqrt(2))/2},{(1/sqrt(2))/2},{2/sqrt(2)});
\coordinate (P3) at ({2.5/sqrt(2)},{1.75/sqrt(2)},{0});
\coordinate (F1) at ({1.3/sqrt(2)},{2.5/sqrt(2)},{2/sqrt(2)});
\shade[ball color = lightgray, opacity = 0.5] (0,0,0) circle (2.0cm);
\tdplotsetrotatedcoords{0}{0}{0};
\draw[dotted, tdplot_rotated_coords, gray] (0,0,0) circle (2);
\tdplotsetrotatedcoords{90}{90}{90};
\draw[dotted, tdplot_rotated_coords, gray] (2,0,0) arc (0:180:2);
\tdplotsetrotatedcoords{0}{90}{90};
\draw[dotted, tdplot_rotated_coords, gray] (2,0,0) arc (0:180:2);
\draw[dotted, gray] (0,0,0) -- (-2,0,0);
\draw[dotted, gray] (0,0,0) -- (0,-2,0);
\draw[dotted, gray] (0,0,0) -- (0,0,-2);
\draw[-stealth] (0,0,0) -- (3.60,0,0) node[below left] {$x$};
\draw[-stealth] (0,0,0) -- (0,2.50,0) node[below right] {$y$};
\draw[-stealth] (0,0,0) -- (0,0,2.70) node[above] {$z$};
\draw[thick, -stealth] (0,0,0) -- (P1) node[above] {$P_1$};
\draw[thick, -stealth] (0,0,0) -- (P2) node[above] {$P_2$};
\draw[thick, -stealth] (0,0,0) -- (P3) node[below] {$P_3$};
\draw[thick, -stealth, blue] (0,0,0) -- (F1) node[above right] {$F$};
\draw[ultra thick, dashed, purple] (F1) -- (P1);
\draw[ultra thick, dashed, purple] (F1) -- (P2);
\draw[ultra thick, dashed, purple] (F1) -- (P3);
\end{tikzpicture}
\label{fig:softmax}
}
\caption{\textbf{All-Distances-Aware Logits, Enhanced Logits, or Logits+.} The illustration presents the difference between IsoMax+ \cite{macedo2021enhanced} and DisMax with respect to \emph{logit formation}. $P_1$, $P_2$, and $P_3$ represent prototypes of classes $1$, $2$, and $3$, respectively. $F$ denotes the feature associated with a given image. Like all current losses, IsoMax+ constructs \emph{each} logit associated with $F$ considering its distance from a \emph{single} prototype (olive dashed line). In contrast, DisMax loss builds \emph{each} logit associated with $F$ considering its distance from \emph{all} prototypes (purple dashed lines). In this paper, we use the terms \emph{all-distances-aware} logits, \emph{enhanced} logits, or logits+ indistinctly.}
\label{fig:logits+}
\end{figure}
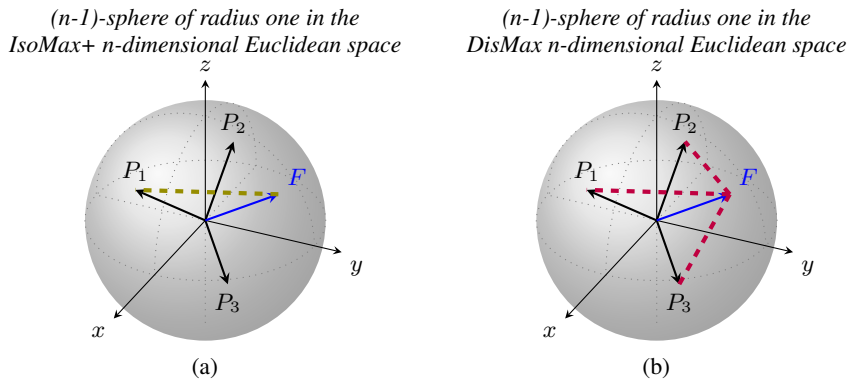

\hl{IsoMax loss replaced the SoftMax loss matrix multiplication with the euclidean distance. It also introduced the entropy maximization trick and showed that it significantly increases the OOD Detection performance. Moreover, it showed that the entropic score improves the OOD detection results compared to the maximum probability score. The IsoMax+ introduced the isometric distance by normalizing the features and the prototypes and adding the learnable distance scale multiplication factor. Finally, it proposed the minimum distance score and showed that it increases the OOD detection performance regarding the entropic score.}

\paragraph{All-Distances-Aware Logits}In IsoMax loss variants (e.g., IsoMax and IsoMax+), logits are formed from distances and are commonly used to calculate the score to perform OOD detection. Hence, it is essential to build logits that contain semantic information relevant to separating in-distribution (ID) from OOD during inference. IsoMax+ uses the \emph{isometric} distances \cite{macedo2021enhanced}. In IsoMax+, the logits are simply the negatives of the isometric distances. We have two motivations to add the \emph{mean} isometric distance considering \emph{all} prototypes to the isometric distance associated with \emph{each} class to construct what we call \emph{all-distances-aware} logits, \emph{enhanced} logits, or logits+.

First, considering that IsoMax+ is an isotropic loss, the pairwise distances between the prototypes and ID examples are forced to become increasingly smaller. Therefore, after training, it is reasonable to believe that ID feature-prototype distances are, on average, smaller than the distances from the prototypes to OOD samples, which were not forced to be closer to the prototypes. Hence, adding the mean distance to the logits used in IsoMax+ can help distinguish between ID and OOD more effectively. Second, taking \emph{all} feature-prototype distances to compose the logits makes them a more stable source of information to perform OOD detection (Fig.~\ref{fig:logits+}).

\begin{equation}
\label{eq:logits+}
\tikzmarknode{x}{\highlight{white}{$L^j_{+}$}}=-\left(\tikzmarknode{y}{\highlight{white}{$D_I^j$}}\:+\tikzmarknode{z}{\highlight{white}{$\frac{1}{N}\sum\limits_{n=1}^N D_I^n$}}\right)
\end{equation}

\begin{equation}
\label{eq:dismax_prob}
\tikzmarknode{x}{\highlight{white}{$\mathcal{P}_{\textsf{DisMax}}(y^{(i)}|\bm{x})$}}=\frac{\exp(E_s\tikzmarknode{z}{\highlight{white}{$L^i_{+}$}}/T)}{\sum\limits_{j=1}^N\exp(E_s\tikzmarknode{y}{\highlight{white}{$L^j_{+}$}}/T)}
\end{equation}

Therefore, we consider an input $\bm{x}$ and a network that performs a transformation $\bm{f}_{\bm{\theta}}(\bm{x})$. We also consider $\bm{p}_{\bm{\phi}}^j$ to be the learnable prototype associated with class $j$. Moreover, considering that $\norm{\bm{v}}$ represents the 2-norm of a vector $\bm{v}$, and $\widehat{\bm{v}}$ represents the 2-norm normalization of $\bm{v}$, we can write the \emph{isometric distance} relative to class $j$ as $D_I^j=\abs{d_s}\:\norm{\widehat{\bm{f}_{\bm{\theta}}(\bm{x})}\!-\!\widehat{\bm{p}_{\bm{\phi}}^j}}$, where $\abs{d_s}$ represents the absolute value of the \emph{learnable} scalar called distance scale \cite{macedo2021enhanced}. Finally, we can write the proposed \emph{enhanced} logit for class $j$ using the equation \eqref{eq:logits+}. $N$ is the number of classes. Probabilities are given by the equation \eqref{eq:dismax_prob}, where $T$ is the temperature. \textcolor{black}{$E_s$ is the entropic scale, which is removed after training \cite{macdo2019isotropic, DBLP:journals/corr/abs-2006.04005, macedo2021enhanced}, but before calibration.} For the rest of this paper, distance means isometric distance.

\paragraph{Fractional Probability Regularization} We often train neural networks using \emph{unitary} probabilities. Indeed, the usual cross-entropy loss forces a \emph{probability equal to one} on a given training example. Consequently, we commonly train neural networks by providing a tiny proportion of points in the learning manifold. Hence, we propose what we call the fractional probability regularization (FPR). The idea is to force the network to learn more diverse points in the learning manifold. Consequently, we confront target and predicted probability distributions also on \emph{fractional} probability values rather than only \emph{unitary} probability manifold points.

\begin{figure}
\centering
\includegraphics[width=\textwidth]{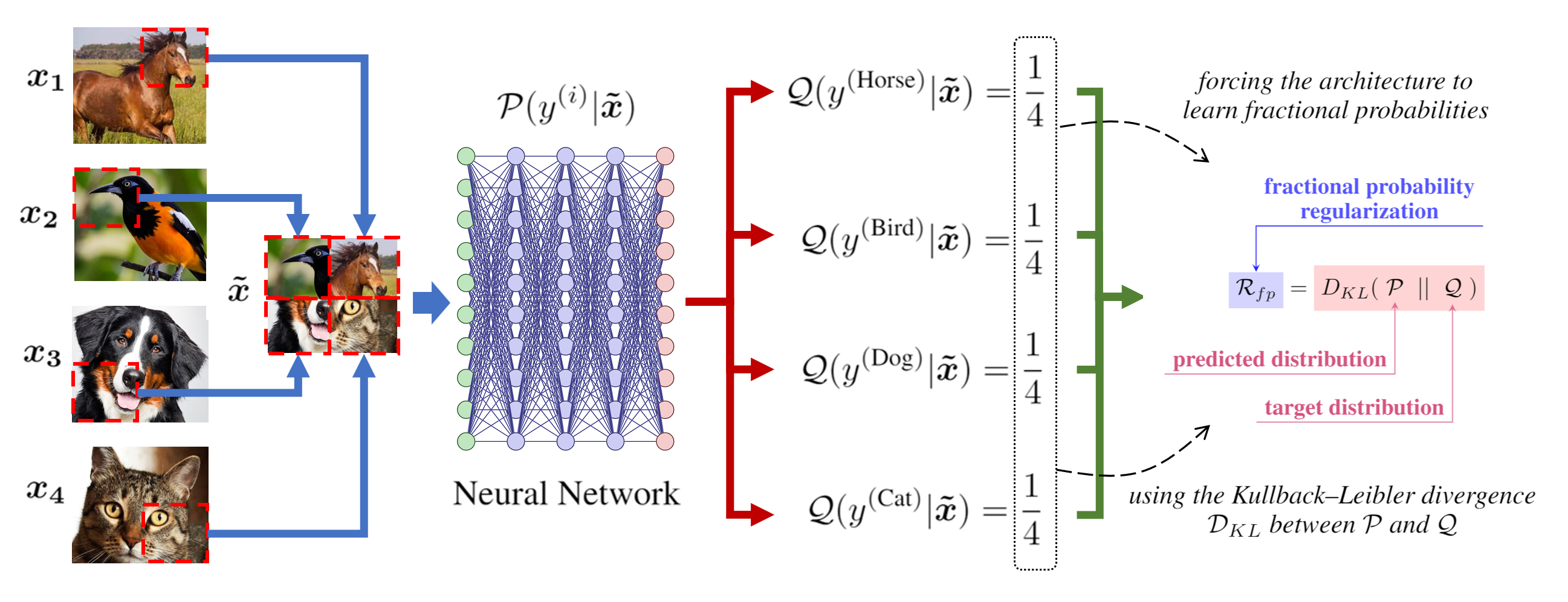}
\caption{\textbf{Fractional Probability Regularization.} We use images composed of patches of four randomly selected training examples. The KL divergence regularization term forces the network to predict fractional probabilities on compound images.}
\label{fig:fpr}
\end{figure}

\begin{align}
\label{eq:fpr}
\tikzmarknode{x}{\highlight{white}{$\mathcal{Q}_{\textsf{Target}}(y^{(i)}|\bm{\tilde{x}})$}}=\tikzmarknode{y}{\highlight{white}{$\frac{1}{4}{\sum\limits_{m=1}^4\delta[y^{(i)}-y^{(j^m)}]}$}}
\end{align}

\begin{equation}
\label{eq:dismax_loss}
\tikzmarknode{w}{\highlight{white}{$\mathcal{L}_{\textsf{DisMax}}$}}=-\log^*\left(\frac{\exp(E_s \tikzmarknode{x}{\highlight{white}{$L^k_{+}$}})}{\sum\limits_j\exp(E_s \tikzmarknode{y}{\highlight{white}{$L^j_{+}$}})}\right) + 
\tikzmarknode{a}{\alpha}\cdot
\tikzmarknode{z}{\highlight{white}{\kld{\mathcal{P}_{\textsf{DisMax}}(y|\bm{\tilde{x}})}{\mathcal{Q}_{\textsf{Target}}(y|\bm{\tilde{x}})}}}
\end{equation}
\blfootnote{\textcolor{black}{\textsuperscript{*}The probability (i.e., the expression between the outermost parentheses) and logarithm operations are computed sequentially and separately for optimal OOD detection performance \cite{macdo2019isotropic}.}}

Therefore, our batch is divided into two halves. In the first half, we use the regular unitary probability training. For the second batch, we construct images specifically composed of patches of four others (Fig. \ref{fig:fpr}). We construct our target probability distribution $Q$ for those images by adding a quarter probability for each class corresponding to a patch of the compound image. Finally, we minimize the KL divergence regularization between our predicted and target probability distributions in the second half. These procedures do not increase training memory requirements. Considering $\delta$ the Kronecker delta function, equation \eqref{eq:fpr} represents the FPR. By combining the enhanced logits and the FPR, equation \eqref{eq:dismax_loss} presents the mathematical expression of the DisMax loss. We always used $\alpha\!=\!1$.

We recognize a similarity between CutMix \cite{DBLP:conf/iccv/YunHCOYC19} and FPR: both are based on the combination of images to create compound data. However, we identify many differences. CutMix combines two images, while FPR combines four images. Moreover, the combination procedure is entirely different. In CutMix, a portion of an image is replaced by a patch of \emph{variable} size, format, and position that comes from another image. In FPR, patches of the \emph{same} size, format, and \emph{predefined} positions from four different images are combined into a single one. This simplification introduced by FPR allowed us to combine four images instead of only two. Indeed, trying to extend CutMix by replacing portions of an image with patches from three others while allowing \emph{random} sizes, shapes, and positions produces \emph{patch superpositions}, making \emph{the calculation of the pairwise ratio of the areas of the superposed patches extremely hard}. Therefore, this simplification made it possible to simultaneously combine four rather than only two images, in addition to avoiding the beta distribution and the related hyperparameter. \textcolor{black}{In FPR, rather than losses, fractional probabilities are proportional to areas.}

While CutMix is applied randomly to some batches with probability~$p$, FPR is applied to half of each batch, avoiding loss or gradient oscillations. CutMix neither creates a target distribution containing fractional probabilities nor forces the predicted probabilities to follow it by minimizing the KL divergence between them. Indeed, CutMix does \emph{not} use the KL divergence at all. CutMix calculates the regular cross-entropy loss of the compound image considering the labels of the original images and takes a linear interpolation between the resulting loss values weighted by the ratio of the areas of the patch and the remaining image. While CutMix operates on losses, FPR operates directly on probabilities \emph{before} calculating loss values. The concept of fractional probabilities is not even present in CutMix. Unlike CutMix, the mentioned procedure can be easily expanded to combine even more than four images. Finally, CutMix \emph{increases the training time} and \emph{presents hyperparameters} \cite{DBLP:conf/iccv/YunHCOYC19}. \hln{Even if we consider adapting the mosaic augmentation from object detection} \cite{DBLP:journals/corr/abs-2004-10934} \hln{to classification and we add to it a CutMix-like regularization, most of the previously mentioned dissimilarities between FPR and CutMix still holds to differentiate FPR from this eventual CutMix-like classification-ported mosaic augmentation.}

\paragraph{Max-Mean Logit Entropy Score}For OOD detection, we propose a score composed of three parts. The first part is the maximum logit+. The second part is the mean logit+. Incorporating the mean value of the logits into the score is an \emph{independent procedure relative to the logit formation}. It can be applied regardless of the type of logit (e.g., usual or enhanced) used during training. Finally, we subtract the entropy calculated considering the probabilities of the neural network output. We call this composite score \emph{Max-Mean Logit Entropy Score} (MMLES). It is given by equation \eqref{eq:mmles}. We call MMLES a \emph{composite} score because it is formed by the sum of many other scores.

\begin{figure}
\centering
\small
\subfloat[]{
\begin{tikzpicture}[line width=1pt, scale=0.6]
\fill[blue!50,even odd rule] (0,0) circle (1.1) (0,0) circle (1.9);
\fill[red!10,even odd rule] (0,0) circle (2.3) (0,0) circle (3.1);
\node[above=0pt of {(0,0)}, outer sep=2pt,fill=white] {ID};
\shade[ball color = black, opacity = 1.0] (0,0) circle (0.1cm);
\shade[ball color = black, opacity = 1.0] (45:1.7) circle (0.1cm);
\shade[ball color = black, opacity = 1.0] (90:1.3) circle (0.1cm);
\shade[ball color = black, opacity = 1.0] (135:1.7) circle (0.1cm);
\shade[ball color = black, opacity = 1.0] (180:1.5) circle (0.1cm);
\shade[ball color = black, opacity = 1.0] (225:1.7) circle (0.1cm);
\shade[ball color = black, opacity = 1.0] (270:1.3) circle (0.1cm);
\shade[ball color = black, opacity = 1.0] (315:1.3) circle (0.1cm);
\foreach \x in {1.5} {\draw [dotted] (0,0) circle (\x cm);}
\foreach \x in {2.7} {\draw [dotted] (0,0) circle (\x cm);}
\draw[dash dot,black,->] (90:1.5) -- (90:3.7) node[above,align=center] {\color{black} \textit{From the point of view of an ID example,}\\\textit{prototypes are likely in the near blue area}};
\draw[->, ultra thick, purple] (0,0) -- (0:1.5);
\end{tikzpicture}
\label{fig:mmles_figa}
}
\hskip 0.5 cm
\subfloat[]{
\begin{tikzpicture}[line width=1pt, scale=0.6]
\fill[blue!10,even odd rule] (0,0) circle (1.1) (0,0) circle (1.9);
\fill[red!50,even odd rule] (0,0) circle (2.3) (0,0) circle (3.1);
\node[above=0pt of {(0,0)}, outer sep=2pt,fill=white] {OOD};
\shade[ball color = black, opacity = 1.0] (0,0) circle (0.1cm);
\shade[ball color = black, opacity = 0.3] (45:1.7) circle (0.1cm);
\shade[ball color = black, opacity = 0.3] (90:1.3) circle (0.1cm);
\shade[ball color = black, opacity = 0.3] (135:1.7) circle (0.1cm);
\shade[ball color = black, opacity = 0.3] (180:1.5) circle (0.1cm);
\shade[ball color = black, opacity = 0.3] (225:1.7) circle (0.1cm);
\shade[ball color = black, opacity = 0.3] (270:1.3) circle (0.1cm);
\shade[ball color = black, opacity = 0.3] (315:1.3) circle (0.1cm);
\shade[ball color = black, opacity = 1.0] (45:2.9) circle (0.1cm);
\shade[ball color = black, opacity = 1.0] (90:2.5) circle (0.1cm);
\shade[ball color = black, opacity = 1.0] (135:2.9) circle (0.1cm);
\shade[ball color = black, opacity = 1.0] (180:2.7) circle (0.1cm);
\shade[ball color = black, opacity = 1.0] (225:2.9) circle (0.1cm);
\shade[ball color = black, opacity = 1.0] (270:2.5) circle (0.1cm);
\shade[ball color = black, opacity = 1.0] (315:2.5) circle (0.1cm);
\draw [->, dashed] (45:1.7) -- (45:2.9);
\draw [->, dashed] (90:1.3) -- (90:2.5);
\draw [->, dashed] (135:1.7) -- (135:2.9);
\draw [->, dashed] (180:1.5) -- (180:2.7);
\draw [->, dashed] (225:1.7) -- (225:2.9);
\draw [->, dashed] (270:1.3) -- (270:2.5);
\draw [->, dashed] (315:1.3) -- (315:2.5);
\foreach \x in {1.5} {\draw [dotted] (0,0) circle (\x cm);}
\foreach \x in {2.7} {\draw [dotted] (0,0) circle (\x cm);}
\draw[dash dot,black,->] (90:2.7) -- (90:3.7) node[above,align=center] {\color{black} \textit{From the point of view of an OOD example,}\\\textit{prototypes are likely in the far red area}};
\draw[->, ultra thick, purple] (0,0) -- (0:2.7);
\end{tikzpicture}
\label{fig:mmles_figb}
}
\caption{\textbf{Max-Mean Logit Entropy Score.} In addition to the maximum logit and the negative entropy, the MMLES incorporates the mean logit+. We empirically observed that the prototypes are generally closer to ID samples than OOD samples, which is true \emph{regardless of whether the ID sample belongs to the class of the considered prototype}. Hence, incorporating this \emph{all-distances-aware} information increases the OOD detection performance (see Table~\ref{tab:dismax_ablation} and Fig.~\ref{fig:histograms}).}
\end{figure}
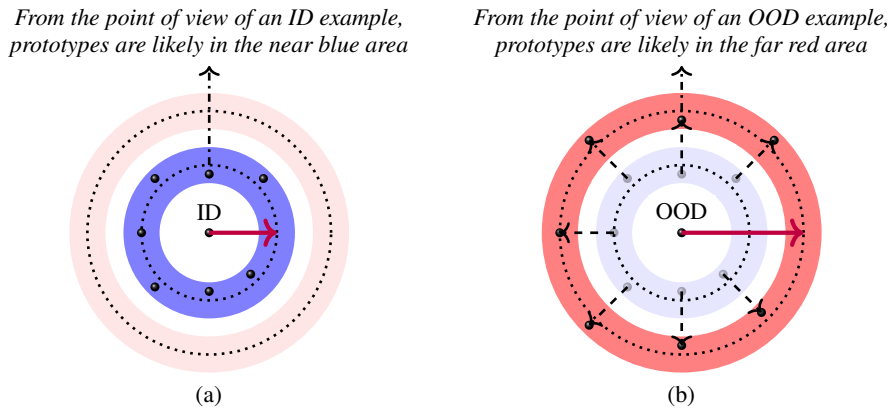

\begin{equation}
\label{eq:mmles}
\tikzmarknode{x}{\highlight{white}{$\mathcal{S}_{\textsf{MMLES}}$}}= \tikzmarknode{w}{\highlight{white}{$\max_j (L^j_{+})$}} + \tikzmarknode{y}{\highlight{white}{$\frac{1}{N}\sum\limits_{n=1}^N L^n_{+}$}} - \tikzmarknode{z}{\highlight{white}{$\mathcal{H(\mathcal{P}_{\textsf{DisMax}})}$}}
\end{equation}

\hl{The mean isometric distance introduced in equation} \eqref{eq:logits+} \hl{is canceled out when computing the temperature-scaled softmax probability in equantion} \eqref{eq:dismax_prob} \hl{from an \emph{analytical} calculus perspective. However, the mean isometric distance affects the logits+ (first term of the above equation) and the mean logits+ (second term of the above equation). Therefore, the mean isometric distance introduced in equation} \eqref{eq:logits+} \hl{surelly also affects the above equation and, consequently, the OOD detection performance (see the ablation study).}

\hl{Additionally, despite not changing the loss from an \emph{analytical} calculus perspective, the addition of the mean isometric distance indeed affects the \emph{training} in the real world of \emph{numeric} computation performed by deep neural networks. This fact is proved by running IsoMax+ and DisMax using the provided code in \emph{deterministic} mode by making the number of execution equal to one. In this case, we clearly observe that running IsoMax+ many times produces the same results. In the same way, running DisMax many times generates the same results. However, the results produced by IsoMax+ and DisMax differ from each other. Not only that, our ablation study shows that DisMax usually outperforms IsoMax+ using the same MDS or MMLES score.}  

\hl{The explanation for the addition of the mean logit+ (second term of the above equation) is provided in Fig.}~\ref{fig:mmles_figa}, \ref{fig:mmles_figb} \hl{and Fig.~}\ref{fig:histograms}. \hl{Finally, the entropy is subtracted because the papers} \cite{macdo2019isotropic,DBLP:journals/corr/abs-2006.04005} \hl{showed that the \emph{negative} of the entropy is a high-quality score for performing OOD detection.} 

\paragraph{Temperature Calibration }Unlike the usual SoftMax loss, the IsoMax loss variants produce \emph{underestimated} probabilities \emph{to obey the Maximum Entropy Principle} \cite{macdo2019isotropic,https://doi.org/10.48550/arxiv.1908.05569,DBLP:journals/corr/abs-2006.04005,macedo2021enhanced}. Therefore, \emph{we need to perform a temperature calibration after training to improve the uncertainty estimation}. To find an optimal temperature\textcolor{black}{, which was kept equal to one during training}, we used the \mbox{L-BFGS-B} algorithm with \emph{approximate} gradients and bounds equal to 0.001 and 100 \cite{doi:10.1137/0916069, 10.1145/279232.279236, 10.1145/2049662.2049669} for ECE minimization \hl{on the validation set}. This calibration takes only a few seconds using the provided code.

\section{Experiments}\label{sec:experiments}

To allow standardized comparison, we used the datasets, training procedures, and metrics that were established in \cite{hendrycks2017baseline} and used in many subsequent papers \cite{liang2018enhancing, lee2018simple, Hein2018WhyRN}. We trained many 100-layer DenseNetBCs with growth rate $k\!=\!12$ (i.e., 0.8M parameters) \cite{Huang2017DenselyNetworks}, 34-layer ResNets \cite{He_2016}, and 28-layer WideResNets (widening factor $k\!=\!10$) \cite{DBLP:conf/bmvc/ZagoruykoK16} on the CIFAR10 \cite{Krizhevsky2009LearningImages} and CIFAR100 \cite{Krizhevsky2009LearningImages} datasets with SoftMax, IsoMax+, and DisMax losses using the same procedures (e.g., initial learning rate, learning rate schedule, weight decay). \hl{We also used TinyImageNet and ImageNet} \cite{Deng2009ImageNetDatabase} \hl{as in-distribution training for 90 with learning rate 0.1 and decay by ten in epochs 30 and 60. We used ImageNet-O as a near and hard out-of-distribuition} \cite{DBLP:conf/cvpr/HendrycksZBSS21}.

We used stochastic gradient descent (SGD) with the Nesterov moment equal to 0.9 with a batch size of 64 and an initial learning rate of 0.1. The weight decay was 0.0001, and we did not use dropout. We trained during 300 epochs. We used a learning rate decay rate equal to ten applied in epoch numbers 150, 200, and 250. 
We used images from TinyImageNet \cite{Deng2009ImageNetDatabase}, the Large-scale Scene UNderstanding dataset (LSUN) \cite{Yu2015LSUNLoop}, CIFAR10, CIFAR100, and SVHN \cite{Netzer2011ReadingLearning} to create out-of-distribution samples. We added these out-of-distribution images to the validation sets of the ID data to form the test sets and evaluate the OOD detection performance. We evaluated the accuracy (ACC) to assess classification performance. Like IsoMax+, when using DisMax, \emph{we only train once}, as no hyperparameter tuning is required.
We evaluated the OOD detection performance using the area under the receiver operating characteristic curve (AUROC), the area under the precision-recall curve (AUPR), and the true negative rate at a 95\% true positive rate (TNR@TPR95). We used the expected calibration error (ECE) \cite{DBLP:conf/aaai/NaeiniCH15,Guo2017OnCO,minderer2021revisiting} for uncertainty estimation performance.

The results are the mean and standard deviation of five runs. Two methods are considered to produce the same performance if their mean performance difference is less than the sum of the error margins.

\begingroup
\begin{table}[!t]
\renewcommand{\arraystretch}{0.9}
\centering
\caption{\textbf{Ablation Study.} MPS means Maximum Probability Score (i.e., the standard for SoftMax loss). MDS indicates Minimum Distance Score (i.e., the standard for IsoMax+ loss). MMLES means Max-Mean Logit Entropy Score (i.e., the standard for DisMax loss for (very) far OOD detection). We used MPS for near OOD detection for DisMax, as this score provided the best results in this particular case. We emphasize that the MPS for DisMax is based on logits+ rather than usual logits. The best performances are bold. All results can be reproduced using the provided code.}
\label{tab:dismax_ablation}
\resizebox{\textwidth}{!}{%
\begin{tabular}{@{}clccccc@{}}
\\
\multicolumn{7}{c}{\Large{CIFAR100}}\\
\toprule
\multirow{5}{*}{Model} & \multirow{5}{*}{Method} & \multirow{3}{*}{\begin{tabular}[c]{@{}c@{}}Score\end{tabular}} & \multicolumn{4}{c}{Out-of-Distribution Detection} \\
\cmidrule{4-7} 
&  & & Near & \multicolumn{2}{c}{Far} & Very Far \\
\cmidrule{4-7} 
&  &  & CIFAR10 & TinyImageNet & LSUN & SVHN \\
\cmidrule{3-7} 
& & MPS,MDS & TNR@95TPR & TNR@95TPR & TNR@95TPR & TNR@95TPR\\
& & MPS/MMLES & (\%) [$\uparrow$] & (\%) [$\uparrow$] & (\%) [$\uparrow$] & (\%) [$\uparrow$]\\
\midrule
\multirow{7}{*}{\shortstack{DenseNetBC100\\(small size)}}
& SoftMax (baseline) \cite{hendrycks2017baseline} & MPS & 17.6$\pm$1.1 & 18.1$\pm$1.7 & 18.7$\pm$2.0 & 19.8$\pm$2.9\\
& IsoMax+ \cite{macedo2021enhanced} & MDS & 17.2$\pm$0.7 & 71.6$\pm$6.5 & 66.8$\pm$9.4 & 67.1$\pm$3.0\\
& \hln{IsoMax+} \cite{macedo2021enhanced} \hln{with CutMix} & MDS & 19.8$\pm$2.1 & 60.8$\pm$9.9 & 57.6$\pm$9.8 & 57.9$\pm$3.5\\
\cmidrule{2-7}

& \multirow{2}{*}{\hln{DisMax (ours)}} & MDS & 15.6$\pm$0.7 & 83.4$\pm$1.6 & 77.0$\pm$3.9 & 84.8$\pm$5.6\\
& & MPS/MMLES & 20.0$\pm$0.7 & 86.6$\pm$1.8 & 80.9$\pm$3.5 & \bf92.9$\pm$2.5\\
\cmidrule{2-7}

& \multirow{2}{*}{DisMax\textsuperscript{$\dagger$} (ours)} & MDS & 16.6$\pm$0.6 & 97.7$\pm$0.3 & 98.5$\pm$0.4 & 57.9$\pm$3.6\\
& & MPS/MMLES & \bf{22.1$\pm$1.1} & \bf{99.0$\pm$0.5} & \bf{99.4$\pm$0.3} & 66.6$\pm$2.6\\
\midrule
\multirow{8}{*}{\shortstack{ResNet34\\(medium size)}}
& SoftMax (baseline) \cite{hendrycks2017baseline} & MPS & 19.4$\pm$0.5 & 20.6$\pm$2.4 & 21.3$\pm$3.4 & 17.1$\pm$5.0\\
& IsoMax+ \cite{macedo2021enhanced} & MDS & 18.0$\pm$0.7 & 43.3$\pm$4.3 & 41.5$\pm$5.7 & 43.6$\pm$3.5\\
& \hln{IsoMax+} \cite{macedo2021enhanced} \hln{with CutMix} & MDS & 18.9$\pm$1.7 & 46.3$\pm$9.6 & 46.5$\pm$9.0 & 35.6$\pm$3.9\\
& \hln{SoftMax with FPR} & MPS & 18.5$\pm$0.5 & 36.4$\pm$6.1 & 37.0$\pm$8.0 & 20.2$\pm$0.9\\
\cmidrule{2-7}

& \multirow{2}{*}{\hln{DisMax (ours)}} & MDS & 18.1$\pm$0.7 & 45.5$\pm$6.9 & 43.6$\pm$7.2 & 53.1$\pm$4.7\\
& & MPS/MMLES & 19.0$\pm$0.9 & 51.7$\pm$6.4 & 47.6$\pm$6.9 & \bf68.3$\pm$6.3\\
\cmidrule{2-7}

& \multirow{2}{*}{DisMax\textsuperscript{$\dagger$} (ours)} & MDS & 20.8$\pm$0.4 & 79.9$\pm$1.5 & 81.5$\pm$1.4 & 43.7$\pm$1.6\\
& & MPS/MMLES & \bf{22.0$\pm$0.5} & \bf{85.4$\pm$1.7} & \bf{86.4$\pm$1.3} & 48.5$\pm$2.0\\
\midrule
\multirow{7}{*}{\shortstack{WideResNet2810\\(big size)}}
& SoftMax (baseline) \cite{hendrycks2017baseline} & MPS & 21.8$\pm$0.7 & 26.7$\pm$5.9 & 28.7$\pm$6.7 & 15.8$\pm$5.5\\
& IsoMax+ \cite{macedo2021enhanced} & MDS & 19.0$\pm$0.7 & 66.9$\pm$3.9 & 67.9$\pm$3.3 & 61.8$\pm$1.9\\
& \hln{IsoMax+} \cite{macedo2021enhanced} \hln{with CutMix} & MDS & 21.5$\pm$1.9 & 52.5$\pm$9.4 & 52.0$\pm$9.1 & 33.3$\pm$9.2\\
\cmidrule{2-7}

& \multirow{2}{*}{\hln{DisMax (ours)}} & MDS & 17.3$\pm$0.9 & 74.4$\pm$5.9 & 74.9$\pm$6.5 & 76.0$\pm$6.3\\
& & MPS/MMLES & 20.7$\pm$0.8 & 83.5$\pm$2.8 & 82.7$\pm$3.8 & \bf91.0$\pm$1.2\\
\cmidrule{2-7}

& \multirow{2}{*}{DisMax\textsuperscript{$\dagger$} (ours)} & MDS & 22.4$\pm$0.2 & 92.3$\pm$1.3 & 95.2$\pm$0.4 & 56.8$\pm$1.8\\
& & MPS/MMLES & \bf{24.6$\pm$0.3} & \bf{96.3$\pm$1.2} & \bf{97.8$\pm$0.9} & 65.6$\pm$1.2\\
\bottomrule
\end{tabular}%
}
\end{table}
\endgroup

\paragraph{Ablation Study}Table \ref{tab:dismax_ablation} shows that logits+ often improve the OOD detection performance compared to IsoMax+. Moreover, it also shows that replacing MDS with the composite score MMLES consistently increases OOD detection. FPR usually increases the OOD detection performance. These conclusions are essentially true regardless of the model, in-distribution, and (near, far, and very far) out-of-distribution. \hl{Finally, we performed experiments combining IsoMax+ with CutMix. However, adding CutMix to IsoMax+ did \emph{not} increase the OOD detection performance significantly. Often, the performance actually \emph{decreases}. Therefore, DisMax\textsuperscript{$\dagger$} easily outperformed IsoMax+ even when the latter was combined with CutMix.}

\begingroup
\setlength{\tabcolsep}{3pt} 
\renewcommand{\arraystretch}{1.2}
\begin{table}
\centering
\caption{\textbf{Classification, Efficiency, Uncertainty, and OOD Detection Results.~}In this table, \mbox{efficiency} represents the inference speed (i.e., the inverse of the inference delay) calculated as a percentage of the performance of a single deterministic neural network trivially trained. For a fair comparison, we also calibrated the temperature of the SoftMax loss and IsoMax+ loss approaches using the same procedure that we defined for DisMax loss. Considering that input preprocessing can be applied indistinctly to improve the OOD detection performance of all methods compared \cite{Hsu2020GeneralizedOD} (at the cost of making their inferences approximately four times less efficient \cite{DBLP:journals/corr/abs-2006.04005}), unless explicitly mentioned otherwise, all results are presented without using input preprocessing. The methods that present the best performances are bold. Results for Scaled Cosine are from Scaled Cosine paper \cite{techapanurak2019hyperparameterfree}. Results for GODIN are from GODIN paper \cite{Hsu2020GeneralizedOD}. Results for Deep Ensemble, DUQ, and SNGP are from SNGP paper \cite{DBLP:conf/nips/LiuLPTBL20}. All other results can be reproduced using the provided code.}
\label{tab:dismax-comparative-results}
\resizebox{\textwidth}{!}{%
\begin{tabular}{@{}clccccccc@{}}
\\
\multicolumn{9}{c}{\Large{CIFAR10}}\\
\toprule
\multirow{5}{*}{Model} & \multirow{5}{*}{Method} & \multirow{3}{*}{Classification} & \multirow{3}{*}{\begin{tabular}[c]{@{}c@{}}Inference\end{tabular}} & \multirow{3}{*}{\begin{tabular}[c]{@{}c@{}}Uncertainty\\ Estimation\end{tabular}} & \multicolumn{4}{c}{Out-of-Distribution Detection} \\ \cmidrule{6-9} 
&  &  &  &  & Near & \multicolumn{2}{c}{Far} & Very Far \\ \cmidrule{6-9} 
&  &  &  &  & CIFAR100 & TinyImageNet & LSUN & SVHN \\ 
\cmidrule{3-9} 
&  & ACC & Efficiency & ECE & AUPR & AUROC & AUROC & AUPR\\
&  & (\%) [$\uparrow$] & (\%) [$\uparrow$] & [$\downarrow$] & (\%) [$\uparrow$] & (\%) [$\uparrow$] & (\%) [$\uparrow$] & (\%) [$\uparrow$]\\
\midrule
\multirow{5}{*}{\shortstack{DenseNetBC100\\(small size)}}
& SoftMax (baseline) \cite{hendrycks2017baseline} & \bf{95.2$\pm$0.1} & \bf100.0 & \bf{0.0043$\pm$0.0008} & 86.2$\pm$0.5 & 92.9$\pm$1.6 & 94.7$\pm$0.9 & 93.7$\pm$3.3\\
& Scaled Cosine \cite{techapanurak2019hyperparameterfree} & \textcolor{black}{94.9$\pm$0.1} & \bf100.0 & - & - & \bf{98.8$\pm$0.3} & \bf{99.2$\pm$0.2} & -\\
& GODIN with preprocessing \cite{Hsu2020GeneralizedOD} & \textcolor{black}{95.0$\pm$0.1} & \textcolor{black}{26.0} & - & - & \bf{99.1$\pm$0.1} & \bf{99.4$\pm$0.1} & -\\
& IsoMax+ \cite{macedo2021enhanced} & \bf{95.1$\pm$0.1} & \bf100.0 & \bf{0.0043$\pm$0.0012} & \bf{90.4$\pm$0.3} & 97.6$\pm$0.9 & 98.3$\pm$0.5 & 99.7$\pm$0.1\\
& DisMax (ours) & \bf{95.1$\pm$0.1} & \bf100.0 & \bf{0.0045$\pm$0.0021} & \bf{90.0$\pm$0.2} & 98.0$\pm$0.5 & 98.4$\pm$0.3 & \bf{99.9$\pm$0.1}\\
\midrule
\multirow{4}{*}{\shortstack{ResNet34\\(medium size)}}
& SoftMax (baseline) \cite{hendrycks2017baseline} & 95.6$\pm$0.1 & \bf100.0 & \bf{0.0060$\pm$0.0013} & 85.3$\pm$0.4 & 89.7$\pm$2.8 & 92.4$\pm$1.6 & 94.9$\pm$1.0\\
& GODIN \cite{Hsu2020GeneralizedOD} & \textcolor{black}{95.1$\pm$0.1} & \bf100.0 & - & - & 95.6$\pm$0.5 & 97.6$\pm$0.2 & -\\
& IsoMax+ \cite{macedo2021enhanced} & 95.5$\pm$0.1 & \bf100.0 & \bf{0.0053$\pm$0.0007} & \bf{90.1$\pm$0.3} & 95.1$\pm$1.0 & 96.9$\pm$0.6 & \bf{98.7$\pm$0.6}\\
& DisMax\textsuperscript{$\dagger$} (ours) & \bf{96.7$\pm$0.2} & \bf100.0 & \bf{0.0058$\pm$0.0008} & \bf{90.3$\pm$0.2} & \bf{98.3$\pm$0.3} & \bf{99.5$\pm$0.1} & \bf{99.1$\pm$0.3}\\
\midrule
\multirow{7}{*}{\shortstack{WideResNet2810\\(big size)}}
& SoftMax (baseline) \cite{hendrycks2017baseline} & 96.2$\pm$0.1 & \bf100.0 & \bf{0.0038$\pm$0.0005} & 87.5$\pm$0.3 & 92.6$\pm$0.9 & 94.0$\pm$0.7 & 95.3$\pm$0.9\\
& Deep Ensemble \cite{lakshminarayanan2017simple} & 96.6$\pm$0.1 & \textcolor{black}{10.3} & 0.0100$\pm$0.0010 & 88.8$\pm$1.0 & - & - & 96.4$\pm$1.0\\
& DUQ \cite{Amersfoort2020SimpleAS} & \textcolor{black}{94.7$\pm$0.1} & \textcolor{black}{45.0} & 0.0340$\pm$0.0020 & 85.4$\pm$1.0 & - & - & 97.3$\pm$1.0\\
& SNGP \cite{DBLP:conf/nips/LiuLPTBL20} & \textcolor{black}{95.9$\pm$0.1} & \textcolor{black}{62.5} & 0.0180$\pm$0.0010 & 90.5$\pm$1.0 & - & - & 99.0$\pm$1.0\\
& Scaled Cosine \cite{techapanurak2019hyperparameterfree} &  \textcolor{black}{95.7$\pm$0.1} & \bf100.0 & - & - & 97.7$\pm$0.7 & 98.6$\pm$0.3 & -\\
& IsoMax+ \cite{macedo2021enhanced} & 96.0$\pm$0.1 & \bf100.0 & \bf{0.0034$\pm$0.0009} & \bf{91.8$\pm$0.1} & 96.6$\pm$0.6 & 97.7$\pm$0.4 & \bf{99.7$\pm$0.3}\\
& DisMax\textsuperscript{$\dagger$} (ours) & \bf{97.0$\pm$0.1} & \bf100.0 & \bf{0.0043$\pm$0.0008} & 90.1$\pm$0.3 & \bf{99.7$\pm$0.1} & \bf{99.9$\pm$0.1} & \bf{99.3$\pm$0.3}\\
\bottomrule
\\
\multicolumn{9}{c}{\Large{CIFAR100}}\\
\toprule
\multirow{5}{*}{Model} & \multirow{5}{*}{Method} & \multirow{3}{*}{Classification} & \multirow{3}{*}{\begin{tabular}[c]{@{}c@{}}Inference\end{tabular}} & \multirow{3}{*}{\begin{tabular}[c]{@{}c@{}}Uncertainty\\ Estimation\end{tabular}} & \multicolumn{4}{c}{Out-of-Distribution Detection} \\ \cmidrule{6-9} 
&  &  &  &  & Near & \multicolumn{2}{c}{Far} & Very Far \\ \cmidrule{6-9} 
&  &  &  &  & CIFAR10 & TinyImageNet & LSUN & SVHN \\
\cmidrule{3-9} 
&  & ACC & Efficiency & ECE & AUPR & AUROC & AUROC & AUPR\\
&  & (\%) [$\uparrow$] & (\%) [$\uparrow$] & [$\downarrow$] & (\%) [$\uparrow$] & (\%) [$\uparrow$] & (\%) [$\uparrow$] & (\%) [$\uparrow$]\\
\midrule
\multirow{5}{*}{\shortstack{DenseNetBC100\\(small size)}}
& SoftMax (baseline) \cite{hendrycks2017baseline} & 77.3$\pm$0.4 & \bf100.0 & 0.0155$\pm$0.0026 & 71.3$\pm$0.8 & 71.8$\pm$2.2 & 73.1$\pm$2.4 & 87.5$\pm$1.5\\
& Scaled Cosine \cite{techapanurak2019hyperparameterfree} & \textcolor{black}{75.7$\pm$0.1} & \bf100.0 & - & - & 97.8$\pm$0.5 & 97.6$\pm$0.8 & -\\
& GODIN with preprocessing \cite{Hsu2020GeneralizedOD} & \textcolor{black}{75.9$\pm$0.1} & \textcolor{black}{24.0} & - & - & 98.6$\pm$0.2 & 98.7$\pm$0.0 & -\\
& IsoMax+ \cite{macedo2021enhanced} & 76.9$\pm$0.3 & \bf100.0 & \bf{0.0108$\pm$0.0017} & 71.3$\pm$0.4 & 95.1$\pm$1.1 & 94.2$\pm$1.7 & \bf{97.4$\pm$0.6}\\
& DisMax\textsuperscript{$\dagger$} (ours) & \bf{79.4$\pm$0.2} & \bf100.0 & 0.0154$\pm$0.0006 & \bf{74.4$\pm$0.2} & \bf{99.8$\pm$0.1} & \bf{99.9$\pm$0.1} & \bf{96.4$\pm$0.8}\\
\midrule
\multirow{5}{*}{\shortstack{ResNet34\\(medium size)}}
& SoftMax (baseline) \cite{hendrycks2017baseline} & 77.7$\pm$0.3 & \bf100.0 & 0.0268$\pm$0.0015 & 73.3$\pm$0.1 & 79.0$\pm$2.1 & 79.6$\pm$1.7 & 86.3$\pm$3.3\\
& GODIN \cite{Hsu2020GeneralizedOD} & \textcolor{black}{75.8$\pm$0.2} & \bf100.0 & - & - & 91.8$\pm$1.1 & 92.0$\pm$0.7 & -\\
& GODIN with dropout \cite{Hsu2020GeneralizedOD} & \textcolor{black}{77.2$\pm$0.1} & \bf100.0 & - & - & \textcolor{black}{87.0$\pm$1.1} & \textcolor{black}{87.0$\pm$2.2} & -\\
& IsoMax+ \cite{macedo2021enhanced} & \textcolor{black}{76.5$\pm$0.3} & \bf100.0 & 0.0190$\pm$0.0025 & 72.1$\pm$0.4 & 89.7$\pm$1.0 & 89.8$\pm$1.3 & \bf{94.5$\pm$0.6}\\
& DisMax\textsuperscript{$\dagger$} (ours) & \bf{80.6$\pm$0.3} & \bf100.0 & \bf{0.0116$\pm$0.0014} & \bf{74.2$\pm$0.6} & \bf{97.6$\pm$0.5} & \bf{97.7$\pm$0.6} & \bf{94.8$\pm$1.0}\\
\midrule
\multirow{7}{*}{\shortstack{WideResNet2810\\(big size)}}
& SoftMax (baseline) \cite{hendrycks2017baseline} & 79.9$\pm$0.2 & \bf100.0 & 0.0272$\pm$0.0032 & 75.4$\pm$0.5 & 81.7$\pm$2.3 & 82.7$\pm$2.2 & 86.0$\pm$2.6\\
& Deep Ensemble \cite{lakshminarayanan2017simple} & 80.2$\pm$0.1 & \textcolor{black}{12.3} & 0.0210$\pm$0.0040 & 78.0$\pm$1.0 & - & - & 88.8$\pm$1.0\\
& DUQ \cite{Amersfoort2020SimpleAS} & \textcolor{black}{78.5$\pm$0.1} & \textcolor{black}{79.9} & 0.1190$\pm$0.0010 & 73.2$\pm$1.0 & - & - & 87.8$\pm$1.0\\
& SNGP \cite{DBLP:conf/nips/LiuLPTBL20} & 79.9$\pm$0.1 & \textcolor{black}{74.9} & 0.0250$\pm$0.0120 & \bf{80.1$\pm$1.0} & - & - & 92.3$\pm$1.0\\
& Scaled Cosine \cite{techapanurak2019hyperparameterfree} & \textcolor{black}{78.5$\pm$0.3} & \bf100.0 & - & - & 95.8$\pm$0.7 & 95.2$\pm$0.8 & -\\
& IsoMax+ \cite{macedo2021enhanced} & \textcolor{black}{79.5$\pm$0.1} & \bf100.0 & 0.0188$\pm$0.0016 & 73.0$\pm$0.8 & 94.2$\pm$2.1 & 94.6$\pm$2.0 & \bf{96.7$\pm$1.7}\\
& DisMax\textsuperscript{$\dagger$} (ours) & \bf{83.0$\pm$0.1} & \bf100.0 & \bf{0.0143$\pm$0.0027} & 76.0$\pm$1.0 & \bf{99.4$\pm$0.2} & \bf{99.6$\pm$0.1} & \bf{97.0$\pm$1.5}\\
\bottomrule
\end{tabular}%
}
\begin{justify}
\scriptsize
We used DisMax for DenseNetBC100 trained on CIFAR10 because this is a very small model and the mentioned dataset has too many examples per class; therefore, no augmentation is needed. For all other experiments, we used DisMax\textsuperscript{$\dagger$}.
\end{justify}
\end{table}
\endgroup

\begingroup
\setlength{\tabcolsep}{3pt} 
\begin{table}
\centering
\caption{\textbf{\hl{Classification, Efficiency, Uncertainty, and OOD Detection Results.}}~\hl{For a fair comparison, we also calibrated the temperature of the SoftMax loss and IsoMax+ loss approaches using the same procedure that we defined for DisMax loss. The methods that present the best performances are bold. All results can be reproduced using the provided code. DisMax variants used MMLES.}}
\label{tab:dismax-comparative-tinyimagenet-results}
\resizebox{\textwidth}{!}{%
\begin{tabular}{@{}clcccccc@{}}
\\
\multicolumn{8}{c}{\large{\hln{TinyImageNet}}}\\
\toprule
\multirow{5}{*}{Model} & \multirow{5}{*}{Method} & \multirow{3}{*}{Classification} & \multirow{3}{*}{\begin{tabular}[c]{@{}c@{}}Uncertainty\\ Estimation\end{tabular}} & \multicolumn{4}{c}{Out-of-Distribution Detection} \\ \cmidrule{5-8} 
&  &  &  & Near & \multicolumn{2}{c}{Far} & Very Far \\ \cmidrule{5-8} 
&  &  &  & \hln{ImageNet-O} & \hln{CIFAR10} & \hln{CIFAR100} & \hln{SVHN} \\ 
\cmidrule{3-8} 
&  & ACC & ECE & AUROC & AUROC & AUROC & AUROC\\
&  & (\%) [$\uparrow$] & [$\downarrow$] & (\%) [$\uparrow$] & (\%) [$\uparrow$] & (\%) [$\uparrow$] & (\%) [$\uparrow$]\\
\midrule
\multirow{4}{*}{\shortstack{\hl{DenseNetBC100\textsuperscript{*}}}}
& SoftMax (baseline) \cite{hendrycks2017baseline} & 61.1$\pm$0.3 & 0.0110$\pm$0.0008 & 64.0$\pm$0.1 & 81.1$\pm$1.3 & 79.6$\pm$0.9 & 84.9$\pm$4.0\\
& IsoMax+ \cite{macedo2021enhanced} & 60.2$\pm$0.3 & \bf{0.0090$\pm$0.0005} & 72.3$\pm$0.7 & 95.8$\pm$1.8 & 93.5$\pm$1.7 & 99.4$\pm$0.2\\
& DisMax (ours) & 60.5$\pm$0.4 & 0.0125$\pm$0.0007 & \bf{78.6$\pm$0.9} & \bf{98.7$\pm$0.3} & \bf{98.1$\pm$0.3} & \bf{99.8$\pm$0.1}\\
& DisMax\textsuperscript{$\dagger$} (ours) & \bf{61.9$\pm$0.2} & 0.0163$\pm$0.0030 & 77.5$\pm$1.0 & 93.2$\pm$2.2 & 93.0$\pm$1.9 & 98.5$\pm$0.5\\
\midrule
\multirow{4}{*}{\shortstack{\hl{ResNet34\textsuperscript{*}}}}
& SoftMax (baseline) \cite{hendrycks2017baseline} & 65.6$\pm$0.3 & 0.0212$\pm$0.0039 & 69.5$\pm$0.4 & 78.3$\pm$1.2 & 76.3$\pm$1.4 & 73.8$\pm$9.7\\
& IsoMax+ \cite{macedo2021enhanced} & 63.7$\pm$0.2 & \bf{0.0149$\pm$0.0024} & 73.9$\pm$0.3 & 90.0$\pm$0.8 & 87.7$\pm$0.5 & 98.0$\pm$0.4\\
& DisMax (ours) & 64.2$\pm$0.3 & \bf{0.0156$\pm$0.0009} & \bf{79.0$\pm$0.2} & \bf{96.9$\pm$0.8} & \bf{95.5$\pm$0.7} & \bf{99.7$\pm$0.1}\\
& DisMax\textsuperscript{$\dagger$} (ours) & \bf{68.0$\pm$0.2} & 0.0260$\pm$0.0027 & 72.8$\pm$0.4 & 83.1$\pm$1.3 & 82.5$\pm$0.7 & 92.9$\pm$5.7\\
\midrule
\multirow{4}{*}{\shortstack{\hl{WideResNet2810\textsuperscript{*}}}}
& SoftMax (baseline) \cite{hendrycks2017baseline} & 67.0$\pm$0.2 & 0.0156$\pm$0.0036 & 68.2$\pm$0.6 & 84.1$\pm$2.4 & 83.3$\pm$2.3 & 90.0$\pm$8.0\\
& IsoMax+ \cite{macedo2021enhanced} & 65.9$\pm$0.3 & \bf{0.0119$\pm$0.0017} & 76.5$\pm$0.6 & 94.6$\pm$0.9 & 92.9$\pm$0.6 & 99.1$\pm$0.6\\
& DisMax (ours) & 65.8$\pm$0.2 & \bf{0.0112$\pm$0.0015} & \bf{81.1$\pm$0.4} & \bf{98.2$\pm$1.0} & \bf{97.6$\pm$0.9} & \bf{99.9$\pm$0.1}\\
& DisMax\textsuperscript{$\dagger$} (ours) & \bf{68.8$\pm$0.1} & 0.0256$\pm$0.0027 & 73.8$\pm$0.7 & 83.6$\pm$3.2 & 83.4$\pm$2.6 & 91.1$\pm$4.0\\
\bottomrule
\end{tabular}%
}
\begin{justify}
\scriptsize
\textsuperscript{*}\textcolor{black}{We used the version of these architectures specially designed for CIFAR10 and CIFAR100. They do not present the same model configuration when training them on ImageNet, in which case a stem is present in the model to reduce the initial resolution of the large-size images.} 
\end{justify}
\footnotesize
\renewcommand{\arraystretch}{0.5}
\begin{tabularx}{\textwidth}{CCC}
\\
\multicolumn{3}{c}{\normalsize{\hln{ResNet18 trained on ImageNet}}}\\
\toprule
Classification & \multicolumn{2}{c}{OOD ImageNet-O} \\ 
ACC (\%) [$\uparrow$] & TNR@95TPR [$\uparrow$] & AUROC (\%) [$\uparrow$] \\ \midrule
\multicolumn{3}{c}{SoftMax / DisMax (ours)} \\ \midrule
69.9 / 69.6 & 1.2 / \bf19.1 & 52.4 / \bf75.8 \\
\bottomrule
\end{tabularx}
\end{table}
\endgroup

\paragraph{Classification, Efficiency, Uncertainty, and OOD Detection Results}Table \ref{tab:dismax-comparative-results} compares DisMax with major approaches such as Scaled Cosine \cite{techapanurak2019hyperparameterfree}, GODIN \cite{Hsu2020GeneralizedOD}, Deep Ensemble \cite{lakshminarayanan2017simple}, DUQ \cite{Amersfoort2020SimpleAS}, and SNGP \cite{DBLP:conf/nips/LiuLPTBL20} regarding classification accuracy, inference efficiency, uncertainty estimation, and (near, far, and very far) out-of-distribution detection. Unlike other approaches, DisMax is as inference efficient as a trivially trained neural network using the usual SoftMax loss. 
Furthermore, DisMax often outperforms other approaches simultaneously in all evaluated metrics.

\hl{Table }\ref{tab:dismax-comparative-tinyimagenet-results}\hl{ present results for TinyImageNet, which has images four times larger than CIFAR datasets. In this case, MMLES always outperformed MPS. DisMax and DisMax\textsuperscript{$\dagger$} work much better than the baseline regarding near OOD. Moreover, with exception to classification accuracy, DisMax outperformed DisMax\textsuperscript{$\dagger$} producing very high performce OOD detection results. Hence, FPR may not be necessary for large-size images.}

\hl{Finally, Table }\ref{tab:dismax-comparative-tinyimagenet-results} \hl{also presents results for the ResNet18 trained on ImageNet. In such a case, DisMax produces similar classification accuracy to a usually trained neural network while significantly increasing the OOD detection performance.}

\begin{figure*}
\centering
\includegraphics[width=\textwidth]{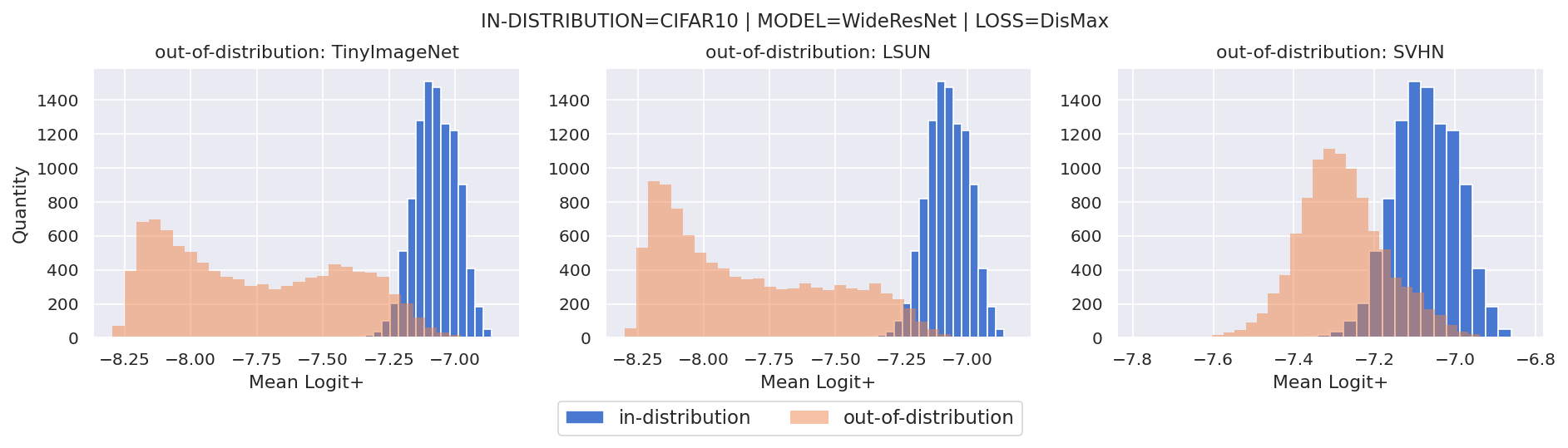}
\caption{\textbf{\textcolor{black}{Max-Mean Logit Entropy Score Analyses.}}~In the feature space, the mean distance from an in-distribution image to \emph{all} prototypes is usually smaller than the mean distance from an out-of-distribution image to the \emph{all} prototypes. For example, consider a given class present in CIFAR10. This figure shows that even prototypes associated with classes \emph{other than the selected class} are usually closer to images of the assumed class (in-distribution in blue) than images that do not belong to CIFAR10 at all (out-of-distributions in orange). This explains why the \emph{mean value} of logits+ considering all prototypes contributes to the OOD detection performance. Therefore, not only the distance to the nearest prototype is used in the mentioned task.}
\label{fig:histograms}
\end{figure*}

\paragraph{\textcolor{black}{Max-Mean Logit Entropy Score Analyses}}Fig.~\ref{fig:histograms} show the distribution of mean logits+ under some scenarios. We see that prototypes are, on average, usually closer to in-distribution examples than out-of-distribution examples, which explains why the  enhanced logit improves OOD detection performance when combined with the maximum logit+ and the negative entropy to compose the MMLES. In other words, even prototypes \emph{that are not associated with the class of a given in-distribution example} are usually closer to it than they are to out-of-distribution examples. 

\paragraph{\textcolor{black}{\hl{Limitations and Societal Impact}}}\hl{We consider requiring a score for near and another to (very) far OOD detection a limitation. Fortunately, for larger images such as the ones from TinyImageNet and ImageNet, the MMLES consistently outperformed the MPS, regardless of considering near or (very) far OOD detection. Moreover, in the TinyImageNet dataset, we observed that FPR does not perform as well as in the CIFAR10 and CIFAR100 datasets, which may indicate that for large-size images DisMax outperforms DisMax\textsuperscript{$\dagger$}. Finally, from a Societal Impact perspective, we may be concerned about how innovations in deep learning may be used, for example, for intrusive tracking.} 

\newpage
\section{Related Works}

In 2019, on the one hand, IsoMax \cite{https://doi.org/10.48550/arxiv.1908.05569} proposed a \emph{non-squared Euclidean distance last layer to address out-of-distribution detection} in an \emph{end-to-end trainable way} (i.e. no feature extraction). On the other hand, Scaled Cosine \cite{techapanurak2019hyperparameterfree} proposed using a \emph{cosine distance}. Although the scale factor in IsoMax is a \emph{constant} scalar called the entropy scale, Scaled Cosine requires the addition of a \emph{block of layers} to learn the scale factor. This is made up of an exponential function, batch normalization, and a linear layer that has the feature layer as input. Moreover, to present high performance, it is necessary to avoid applying weight decay to this \emph{extra learning block}. We believe that this additional learning block, which adds an ad hoc linear layer in the final of the neural network, may make the solution prone to \emph{overfitting} and explain the classification accuracy drop mentioned by the authors.

In 2020, GODIN \cite{Hsu2020GeneralizedOD} cited and was inspired by Scaled Cosine. GODIN kept the \emph{extra learning block} to learn the scale factor and also avoided applying weight decay to it. In addition to the usual affine transformation and cosine distance from Scaled Cosine, it presents a variant that uses a Euclidean distance-based last layer, similar to IsoMax. The major contribution of GODIN was to allow using the input preprocessing introduced in ODIN without the need for out-of-distribution data. However, input preprocessing increases the inference latency (i.e., reduces the inference efficiency) approximately four times \cite{DBLP:journals/corr/abs-2006.04005}. Moreover, SNGP \cite{DBLP:conf/nips/LiuLPTBL20} cited, followed, and improved the idea introduced by IsoMax in 2019: A \emph{distance-based output layer} for OOD detection. In a similar direction, DUQ \cite{Amersfoort2020SimpleAS} also proposed a modified \emph{distance-based loss to address OOD detection}. However, unlike IsoMax variants (e.g., IsoMax, IsoMax+, and DisMax), SNGP and DUQ produce inferences not as efficient as those produced by a deterministic neural network \cite{DBLP:conf/nips/LiuLPTBL20}. Moreover, they require training the neural network many times for hyperparameter tuning. \textcolor{black}{In 2021, the IsoMax+ \cite{macedo2021enhanced} introduced the \emph{isometric} distance.}

\section{Conclusion}

In this work, we proposed DisMax by improving the IsoMax+ with the \emph{enhanced} logits and the \emph{Fractional Probability Regularization}. We also presented a novel \emph{composite} score called MMLES for OOD detection by combining the maximum logit+, the \emph{mean} logit+, and the negative entropy of the network output. We proposed a \emph{simple and fast temperature scaling procedure} performed after training that makes DisMax produce a high-performance uncertainty estimation. Our experiments showed that the proposed method commonly outperforms the current approaches simultaneously in classification accuracy, inference efficiency, uncertainty estimation, and out-of-distribution detection.

\newpage

\bibliographystyle{icml2021}
\bibliography{references}

\end{document}